# Contrast-Free Autonomous Navigation of Untethered Endovascular Microrobots Using Single-Plane Fluoroscopy

Husnu Halid Alabay[1,2], Tuan-Anh Le[1], Ping Wang[1], Hakan Ceylan[1,2]*

[1] Department of Physiology and Biomedical Engineering, Mayo Clinic, Scottsdale, Arizona, USA

[2] School of Biological and Health Systems Engineering, Arizona State University, Tempe, AZ, USA

*Corresponding author: Hakan Ceylan (Ceylan.Hakan@mayo.edu)

## ABSTRACT

Reliable three-dimensional (3D) navigation of magnetically actuated untethered microrobots remains a major barrier to clinical translation. X-ray fluoroscopy is the standard real-time imaging modality for endovascular procedures, but single-plane fluoroscopy provides only a two-dimensional (2D) projection, eliminating depth information and complicating autonomous navigation. Recovering this information through biplane imaging or repeated contrast-enhanced angiography increases procedural complexity, radiation exposure, or contrast burden. Here, we introduce VISTA (Virtual Integration for Spatial Tracking and Autonomy), a digital twin framework enabling contrast-free autonomous navigation under single-plane fluoroscopy. VISTA reconstructs vascular anatomy as a 3D digital twin, discretizes the vessel centerline into navigation milestones, and assigns the detected 2D robot position to the nearest projected milestone. Consecutive milestones define the local vessel orientation used to generate magnetic actuation commands, converting single-plane fluoroscopic observations into topology-constrained navigation states without requiring contrast injection during navigation. VISTA is demonstrated across anatomically distinct vascular phantoms under continuous flow and within the inferior vena cava of a live rat in vivo. Compared with conventional fluoroscopic human-in-the-loop control, VISTA reduced navigation time by up to 62%, corrective actuation commands by up to 98%, and radiation exposure by up to 57%. These results establish VISTA as a digital twin-guided framework for contrast-free autonomous navigation of untethered endovascular microrobots using widely available single-plane fluoroscopy.

**Keywords:** Digital twin; Fluoroscopic imaging; Magnetic actuation; Autonomous navigation; Millirobot; Endovascular robotics

## INTRODUCTION

Untethered endovascular microrobots offer a fundamentally different approach to minimally invasive intervention. Their small size, wireless actuation, and passive compliance enable access to anatomically tortuous vasculature, including regions with fragile or inflamed vessel walls, where conventional tethered devices cannot reach without mechanical trauma [1-8]. These properties position such systems to expand endovascular therapy toward more precise, atraumatic treatment across a broader range of cardiovascular and neurovascular conditions [9, 10]. Realizing this potential, however, depends critically on the ability to localize, track, and navigate untethered robots reliably under clinically feasible conditions.

Magnetic actuation has been widely adopted for remote control of endovascular microrobots, enabling torque- and force-based manipulation without physical tethers [11-13]. Effective magnetic navigation requires continuous estimation of the robot's 3D position and orientation in order to update the external magnetic field in real time. In laboratory settings, stereo optical imaging triangulates robot pose effectively [14]. In vivo, optical line-of-sight is physically inaccessible, as biological tissue opacity eliminates this option entirely, necessitating imaging modalities compatible with clinical endovascular workflows.

X-ray fluoroscopy is the clinical standard for real-time image guidance during endovascular procedures, including angiography, balloon angioplasty, stenting, and embolization [15-17]. Fluoroscopy readily visualizes radiopaque interventional devices but does not inherently delineate the surrounding vessel lumen. Vascular anatomy is therefore conventionally visualized through contrast-enhanced angiography, with additional contrast-enhanced angiographic acquisitions performed as needed during device navigation to re-establish vascular context [18]. Although this approach is well established for catheter-based interventions, repeated contrast administration increases the cumulative procedural contrast burden, making contrast-sparing strategies particularly desirable in patients at increased risk of contrast-associated renal complications [19, 20].

This limitation becomes more consequential for untethered microrobots. Conventional catheters remain mechanically tethered and can be directly advanced, retracted, or repositioned by the operator. Once released into the circulation, an untethered robot lacks this direct mechanical control and must instead be localized and actuated remotely. Reliable navigation therefore requires knowledge of both the robot location and the surrounding vascular trajectory. Repeated contrast-enhanced angiography could intermittently recover this vascular context, but only transiently, requiring additional contrast administration whenever the vessel trajectory must be re-visualized. A navigation framework that retains vascular information throughout the procedure could instead provide persistent anatomical context without repeatedly administering contrast. Errors in localization or magnetic-field orientation may otherwise result in stalling, unintended downstream migration, or incorrect branch selection.

A second limitation is inherent to the imaging geometry itself. Single-plane fluoroscopy collapses 3D vascular anatomy onto a 2D projection, eliminating depth and obscuring local vessel orientation and branching context [21, 22]. A microrobot may therefore be clearly visible fluoroscopically while its position along the 3D vessel trajectory remains ambiguous **(Fig. 1A)**. In complex geometries, multiple distinct 3D locations can produce similar or overlapping 2D projections, directly complicating closed-loop magnetic navigation [23]. Biplane fluoroscopy can partially recover depth by providing simultaneous projections from two imaging orientations, but introduces additional radiation exposure, calibration requirements, and potential workspace conflicts with external magnetic actuation hardware [24-27]. Moreover, biplane systems are not universally available, limiting their scalability across clinical environments [9]. An alternative approach is therefore needed that extracts sufficient spatial information for autonomous navigation from widely available single-plane fluoroscopy without requiring repeated contrast administration.

These limitations also increase the demands placed on manual magnetic control. Operators must interpret 2D fluoroscopic projections, infer motion relative to the 3D vascular trajectory, manage contrast administration, and simultaneously adjust the external magnetic actuator under dynamic flow [28]. Because untethered magnetic actuation lacks direct mechanical coupling to the device, errors in field orientation or timing can result in off-target motion, migration, or deformation. Operator reaction time and fatigue may therefore limit the consistency of purely manual fluoroscopy-guided magnetic navigation in complex vascular geometries [29, 30].

Prior approaches to endovascular microrobot navigation have incorporated flow-aware planning and model-based steering [31], but experimental validation has largely been restricted to planar geometries, optically accessible environments, or non-clinical imaging modalities. Fluoroscopy-guided studies in vascular phantoms have similarly not addressed the depth ambiguity introduced by single-plane projection imaging in anatomically realistic 3D vascular networks with overlapping vessel trajectories [32]. Consequently, autonomous navigation using single-plane fluoroscopy remains limited by the absence of persistent 3D vascular context.

We previously introduced an X-ray-guided magnetic manipulation platform combining fluoroscopic imaging with a virtual workspace and robot avatar for human-in-the-loop 2D steering [33]. That platform achieved low-latency fluoroscopic control but did not resolve depth ambiguity or support autonomous navigation through 3D vascular geometries. In related work, we developed EndoBot, a physiologically adaptive soft magnetic microrobot for atraumatic image-guided endovascular therapy [34]. Together, these developments motivated a navigation framework that could combine real-time fluoroscopic robot detection with pre-existing 3D vascular information to enable autonomous magnetic control.

Here, we present Virtual Integration for Spatial Tracking and Autonomy (VISTA), a digital twin framework for contrast-free autonomous navigation of untethered endovascular microrobots using single-plane fluoroscopy. VISTA represents the known 3D vascular centerline as a sequence of discrete navigation milestones and associates each 2D fluoroscopic robot detection with the corresponding state along this predefined vascular trajectory. Consecutive milestones provide the local vessel orientation used to configure the rotating magnetic field for closed-loop actuation **(Fig. 1B)**. Because the vascular topology is encoded within the digital twin, fluoroscopy is used to track the radiopaque robot rather than repeatedly visualize the vessel lumen, eliminating the need for contrast administration during navigation. Rather than attempting unconstrained reconstruction of the robot's 3D pose from a single projection, VISTA leverages known vascular topology to recover the spatial information required for milestone-to-milestone autonomous control.

We validate VISTA using our helical soft microrobot, EndoBot, across diverse anatomically distinct vascular phantoms under continuous flow and within a live rat inferior vena cava model in vivo, demonstrating autonomous closed-loop navigation with reduced operator intervention, lower cognitive burden, shorter navigation times, lower radiation exposure, and fewer corrective magnetic actuation commands compared with conventional fluoroscopic human-in-the-loop control. By transforming single-plane fluoroscopy from a passive visualization modality into an active input for closed-loop navigation, VISTA establishes a scalable digital twin-guided framework for clinically deployable untethered endovascular microrobots.

## RESULTS

### *Centerline Milestones Define the Navigation State*

A patient-specific vascular digital twin was reconstructed in the Unity environment using the segmented 3D angiographic model of the target vascular segment and its corresponding projection images **(Fig. 2A)**. The reconstructed vascular mesh was skeletonized to extract the vessel centerline, represented as a spatial curve $\boldsymbol{c}(s)$, parameterized by arc length $s \in [0, L]$. The centerline was then discretized into an ordered set of navigation milestones,

$$\mathcal{M} = \{\boldsymbol{m}_i = c(s_i)\}_{i=1}^{N},$$

where $\boldsymbol{m}_i \in \mathbb{R}^3$denotes the 3D coordinates of the $i$-th milestone [35]. These milestones define the discrete navigation states along the vascular topology. At each fluoroscopic frame $t$, EndoBot was detected at the image-plane coordinate

$$\boldsymbol{u}_t = [u_t, v_t]^\top \in \mathbb{R}^2$$

where $u_t$ and $v_t$ denote the horizontal and vertical pixel coordinates of the detected EndoBot centroid, respectively. The *current navigation state* was assigned to the milestone whose projected image-plane location was closest to the detected robot position,

$$i_t^* = \arg\min_i \|\boldsymbol{u}_t - \Pi(\boldsymbol{m}_i)\|_2,$$

where $\Pi(\cdot)$ denotes the projection operator that maps a 3D point in the registered digital twin to its corresponding location on the fluoroscopic image plane. The local navigation direction was then defined by the vector connecting the previous and current navigation milestones,

$$\hat{\boldsymbol{t}}_{i_t^*} = \frac{\boldsymbol{m}_{i_t^*} - \boldsymbol{m}_{i_t^*-1}}{\left\|\boldsymbol{m}_{i_t^*} - \boldsymbol{m}_{i_t^*-1}\right\|_2},$$

which approximates the local vessel tangent over the current vessel segment.

The active milestone pair also defines the rotation axis of the commanded magnetic field. Accordingly, the magnetic actuation command for milestone $i$ is

$$\mathbf{B}_i(\tau) = B_0 R_{\hat{t}_i}(\omega\tau)\mathbf{b}_{0,i},$$

where $B_0$ is the prescribed magnetic-field magnitude, $\omega$ is the rotation frequency, $b_{0,i}$ is an initial unit vector satisfying $b_{0,i}^\top \hat{t}_i = 0$, and $R_{\hat{t}_i}(\omega\tau)$ denotes rotation about the local milestone axis. Consequently, the magnetic field rotates within the plane orthogonal to the local vessel orientation, generating the magnetic torque required for helical propulsion.

Rather than continuously recomputing the magnetic-field rotation axis from a continuously estimated robot pose, VISTA updates the commanded rotation axis only when the nearest-milestone index changes. Therefore,

$$\hat{\mathbf{t}}_{\mathrm{cmd}}(\tau) = \hat{\mathbf{t}}_{i_t^*},$$

where $i_t^*$ remains constant between milestone transitions. Autonomous navigation is therefore formulated as an event-driven process constrained by the known vascular topology **(Fig. 2B)**.

This event-driven update strategy reduces unnecessary robot-arm motion. In principle, the vessel centerline could be represented as a continuously sampled trajectory with correspondingly frequent updates of the magnetic-field rotation axis. During preliminary hardware testing, however, frequent orientation updates generated rapid manipulator commands that repeatedly triggered robot-controller safety limits and prevented reliable locomotion. This limitation was amplified by the kinematic relationship between the EndoBot and the external magnet, whereby relatively small changes in the desired magnetic rotation axis could require comparatively large Cartesian motions of the robot arm.

The useful spatial resolution of this control loop is further bounded by both the manipulator and fluoroscopic imaging system. Robot-arm commands below the minimum effective end-effector displacement (~0.1 mm) do not produce meaningful changes in magnet pose, while EndoBot displacements below the fluoroscopic spatial resolution (~0.3 mm) cannot be reliably resolved for subsequent localization updates [36]. Increasing navigation-state resolution beyond these limits therefore provides little additional control information while increasing command frequency.

Center spline creation is also bounded by voxelization resolution where a smaller number of voxel cubes cannot cover the target phantom while higher number of voxel cubes creates unnecessary calculation burdens. Also, milestone spacing consequently represents a trade-off between geometric fidelity and control efficiency. Smaller spacing more closely approximates local vessel geometry but increases navigation-state transitions and manipulator reorientation, whereas larger spacing permits longer periods of uninterrupted magnetic propulsion at the expense of local directional accuracy, particularly in highly curved vessel segments. VISTA therefore selects milestone spacing to preserve the local vessel orientation required for effective magnetic actuation while avoiding unnecessarily frequent manipulator updates (Supplementary Text 1).

Because the vessel centerline varies continuously while the commanded magnetic-field rotation axis remains piecewise constant between milestone transitions, milestone-based navigation introduces a transient angular mismatch between the commanded actuation axis and the instantaneous local vessel orientation. Each orientation update occurs only after the EndoBot passes the midpoint between neighboring milestones, producing a maximum angular mismatch within each milestone interval **(Fig. 2C)**. This mismatch alters the effective magnetic torque and propulsion direction and therefore influences navigation performance.

To characterize this effect, complementary in vitro experiments and COMSOL simulations were performed. Experimentally, the magnetic-field rotation axis was intentionally offset from the EndoBot orientation from 0° to 90° in 15° increments. EndoBot forward displacement was measured fluoroscopically at each angular offset while maintaining a 90-mm separation between the EndoBot and magnet center. This separation provided the operating condition used for magnetic actuation, and the maximum measurable forward displacement was limited to 25 mm, corresponding to the distance from the magnet center to its edge. Forward displacement progressively decreased with increasing angular mismatch, confirming that orientation error reduces propulsion performance.

In parallel, COMSOL simulations quantified the corresponding changes in magnetic-field magnitude and magnetic torque as a function of angular mismatch. To define practical actuation limits, 95% and 90% retention of the nominal magnetic response were selected as engineering thresholds. The 95% retention criterion corresponded to angular deviations of 18.33° and 18.20° for magnetic field and torque, respectively, whereas the 90% criterion corresponded to 25.74° and 25.85°. Averaging the field- and torque-derived values yielded angular tolerances of 18.26° and 25.79° for 95% and 90% magnetic-actuation retention, respectively.

We next determined how milestone spacing translates into angular mismatch across the vascular geometries used in this study. The centerlines of nine phantom geometries used in this study were discretized using milestone spacings of 1, 2, 5, 10, and 20 mm, and the maximum angular deviation between the piecewise-constant actuation axis and local vessel orientation was determined for each geometry and spacing. Across all nine geometries, the worst-case angular deviations were 7.93°, 8.57°, 13.59°, 22.01°, and 34.90° for milestone spacings of 1, 2, 5, 10, and 20 mm, respectively **(Fig. 2D)**.

Finally, the actuation-derived angular tolerances were mapped onto this geometry-derived relationship. Linear interpolation yielded maximum milestone spacings of 7.77 mm for ≥95% magnetic-actuation retention and 12.93 mm for ≥ 90% retention **(Fig. 2E)**. Thus, milestone spacing can be selected directly from the allowable loss in magnetic actuation, providing a quantitative criterion that links vascular geometry, magnetic actuation, and navigation-state discretization. Together, these constraints establish an operational range for milestone placement: the lower bound is governed by imaging and control resolution, whereas the upper bound is determined by the allowable reduction in magnetic actuation. Selecting milestone spacing within these limits enables sufficiently resolved EndoBot localization while maintaining approximately 90-95% of the desired magnetic actuation performance and suppressing unnecessary robot-arm motion.

### *Accurate Detection, Localization and Real-Time Position Estimation*

Before autonomous navigation, each vascular phantom was reconstructed using the 3D acquisition mode of the fluoroscopy system, in which the imaging system rotated around the target to acquire projections for volumetric reconstruction. The reconstructed volume was visualized in frontal, lateral, and axial views and incorporated into Unity as a vessel-specific digital twin. Initial correspondence between the fluoroscopic image and virtual model was established using a common image origin and matched horizontal and vertical coordinates. X-ray-visible fiducial markers were subsequently used to align the physical phantom and digital twin in position, scale, longitudinal orientation, and x-axis rotation. Comparisons across RGB imaging, fluoroscopy, Fusion 360, and Unity confirmed spatial consistency across the physical and virtual environments (Supplementary Text 2).

To validate milestone-based localization, prerecorded fluoroscopic sequences acquired during EndoBot navigation were processed by VISTA (**Fig. 3A**). In each frame, the EndoBot detector provided its 2D image-plane position, which was matched to the nearest projected centerline milestone. The corresponding milestone provided the discrete 3D navigation state within the registered digital twin. The local orientation associated with that navigation state was subsequently obtained from the active milestone pair and used to define the magnetic-field rotation axis (**Fig. 3B**).

Detection robustness was evaluated independently across varying EndoBot orientations using an additional vascular phantom imaged at 15° increments from 0° to 90°. The EndoBot-specific detection model maintained > 90% detection accuracy at every tested orientation and achieved an overall detection rate of 99.2%. Unlike bead-based detectors with limited orientation-dependent visibility, a single EndoBot detector reliably identified the robot throughout the tested angular range, supporting localization across the orientations encountered during 3D vascular navigation.

***Real-Time Robotic Execution and Path-Following Validation***

VISTA's robotic execution capabilities were first characterized independently to verify accurate translation, rotation, and command delivery. Unity transmitted Cartesian position and quaternion-based orientation commands to the KUKA robotic arm through the Fast Robot Interface. Linear interpolation and spherical linear interpolation were used to generate smooth transitions between consecutive position and orientation commands. The manipulator accurately executed prescribed translations of 10-30 cm and rotations of 0°-90°, with positional and angular errors below 0.5 mm and 0.5°, respectively.

We next characterized the complete perception-to-command pipeline to determine whether each control update could be generated within the fluoroscopic imaging interval. The measured pipeline included fluoroscopic image acquisition, EndoBot detection, milestone-based localization, ROS2 communication, Unity-based navigation-state processing, and robot-arm command generation. End-to-end latency was approximately 20 ms, allowing each localization and control cycle to be completed before acquisition of the subsequent fluoroscopic frame.

Following these hardware and timing validations, the complete milestone-based control strategy was evaluated using a sinusoidal virtual vessel trajectory containing 200 centerline points and 17 navigation milestones generated in Python **(Fig. 3C)**. As the virtual EndoBot progressed along the trajectory, each milestone transition generated an updated local rotation axis and corresponding magnetic actuation command. The robotic manipulator sequentially repositioned and reoriented the external magnet according to these commands. Close agreement between commanded and executed poses demonstrated accurate and repeatable translation of VISTA-generated navigation states into physical robotic motion.

***VISTA Generalizes Across Vascular Anatomies***

After validation of digital-twin reconstruction, spatial registration, robotic command execution, and real-time processing, the geometric robustness of VISTA was evaluated in three in vitro vascular phantoms reconstructed from human umbilical veins **(Fig. 4**). The phantoms represented distinct vascular geometries with different curvature profiles and luminal diameters.

For each phantom, a vessel-specific digital twin and corresponding centerline milestone map were generated without modification to the underlying navigation algorithm or control parameters. The EndoBot successfully navigated from the inlet to the outlet under fully autonomous closed-loop control in all three phantoms. These experiments demonstrate that VISTA can accommodate anatomically distinct vascular geometries through the generation of geometry-specific digital twins and navigation maps rather than through geometry-specific controller tuning.

***Autonomous Closed-Loop Navigation Under Blood Flow***

Integrated in vitro testing combining the C-arm, KUKA robotic arm, peristaltic pump, and vascular phantom validated fully autonomous closed-loop navigation under controlled porcine-blood flow **(Fig. 5A-i-ii)**. Whole porcine blood was circulated through the 2.1-mm-inner-diameter lumen using a peristaltic pump (Longer G100-2J pump equipped with a DG15-24 pump head) at 3.0 rpm where its equivalent flow rate (~2.5 mL/min) and lumen velocity (11.8 mm/s) estimated from manufacturer-provided water calibration. EndoBot first navigated from left to right in the direction of flow, completing the approximately 162-mm centerline path in 15 seconds using nine unique robot-arm movements **(Fig. 5A iii-iv)**. It then reversed direction through opposite external magnetic field rotation and returned to its starting point by going against the flow in 29 seconds using 15 root arm movements, without human intervention where the Unity interface additionally permitted navigation to be paused and resumed during magnet reorientation, providing a modular control architecture compatible **(Fig. 5A-iii-vi)**. The resulting EndoBot velocities were approximately 10.8 mm/s with the flow and 5.59 mm/s against the flow. Thus, upstream navigation required 93.3% more time and 66.7% more robot-arm movements, while reducing the mean traversal velocity by 48.3% and the centerline advancement per movement from approximately 18.0 to 10.8 mm. These directional differences indicate that opposing flow increased the number of incremental actuation and trajectory-correction steps required for successful navigation.

### *Autonomous Closed-Loop Navigation in the Inferior Vena Cava of an In Vivo Rat Model*

VISTA was further validated through an in vivo experiment in a rat model using the same integrated imaging, robotic actuation, and autonomous control platform **(Fig. 5B-i**). Following preparation of the animal for surgery, the inferior vena cava (IVC) was surgically exposed. The EndoBot, measuring 1.8 mm in diameter, was loaded into a catheter sheath and introduced into the IVC as described in our previous work [37]. After the EndoBot was manually deployed from the catheter sheath, the autonomous navigation system was initiated.

The target-detection region and the corresponding Unity-based vascular model were then configured, and communication between the imaging, simulation, and control components was established **(Fig. 5B-ii)**. A predefined green region of interest was applied to the fluoroscopic image, and only detections located within this region were considered valid EndoBot candidates. Each candidate detection was processed according to its confidence score, and the detection with the highest confidence was selected and published through ROS2. Restricting the detection process to this region reduced false-positive detections caused by surgical instruments, anatomical structures, and other objects visible within the fluoroscopic field.

During the experiment, the EndoBot was autonomously navigated within the IVC between the catheter tip and the level of the diaphragm **(Fig. 5B-iii-iv)**. Torque-based magnetic actuation remained the primary mechanism for EndoBot propulsion; however, under certain in vivo conditions, rotational actuation alone was insufficient to produce consistent translational motion, potentially due to local blood flow and animal-specific vascular geometry. To improve propulsion under these conditions, the control strategy was augmented with a margin-based translational adjustment. When the EndoBot remained stationary despite the commanded magnetic orientation, an additional translational offset was applied to the robot arm in the intended direction of EndoBot motion. This adjustment increased the magnetic gradient force acting on the EndoBot, providing additional assistance to facilitate forward or backward propulsion. A translational margin of 25 mm was selected to enhance propulsion while remaining within the operating range of the permanent-magnet actuation system.

### *Autonomous Navigation Outperforms Human Fluoroscopic Control*

To evaluate the performance benefits of autonomous control, VISTA was benchmarked against manual control under fluoroscopic guidance in a 3D helical vascular phantom requiring multiplanar navigation **(Fig. 6A)**. The phantom was mounted on a custom 3D-printed support structure and incorporated fiducial markers to enable 3D registration **(Fig. 6B)**. A novice operator with approximately 10 h of experience, an expert operator with approximately 90 h of experience, and VISTA each completed three trials under standardized conditions. During each trial, navigation was performed along the vascular curvature while maintaining the prescribed magnet–EndoBot distance and relying exclusively on fluoroscopic feedback. All trials were recorded and subsequently evaluated using frame-by-frame analysis **(Fig. 6C)**.

On average, the novice operator required 213.7 s, 120.3 recorded movement actions, and 2.35 mGy of X-ray exposure. The expert operator required 152.3 s, 101.7 movement actions, and 1.93 mGy, whereas VISTA completed the task in 80.7 s using 2.67 movement actions and 1.02 mGy. Relative to the novice condition, VISTA reduced navigation time, movement count, and X-ray exposure by 62.2%, 97.8%, and 56.7%, respectively. Relative to the expert condition, the corresponding reductions were 47.0%, 97.4%, and 47.2%. Overall differences among the three control conditions were significant for total navigation time, X-ray exposure, and movement count. Exploratory directional comparisons also indicated significantly lower total values for VISTA than for both manual conditions. Direction-specific analyses showed significant VISTA improvements during forward propulsion for all three outcomes relative to both operators. During backward propulsion, VISTA significantly reduced time and exposure relative to the novice operator and reduced movement count relative to both manual conditions. Additional tests were performed under non-standardized conditions, including improper magnetic-field alignment with vessel curvature and increased rotational speed. Under improper curvature alignment, novice performance deteriorated and the expert operator was unable to complete the navigation due to EndoBot deformation, whereas VISTA maintained successful navigation with lower time, movement count, and X-ray exposure. Increasing rotational speed provided no clear benefit during manual control, but VISTA reduced navigation time from 80.7 to 66 s and X-ray exposure from 1.02 to 0.63 mGy. Additional tests in a less tortuous phantom showed that these effects were geometry-dependent, highlighting the importance of curvature-aware field alignment and controlled rotational actuation.

## DISCUSSION

This study demonstrates that known vascular topology can be used to overcome key spatial limitations of single-plane fluoroscopy for autonomous navigation of untethered endovascular microrobots. Rather than attempting unconstrained reconstruction of 3D robot pose from a single 2D projection, VISTA identifies anatomically permissible navigation states along a registered vascular centerline. The milestone representation simultaneously provides the spatial context required for localization and the local vessel orientation required for magnetic actuation. By coupling these functions within a digital twin, VISTA enables closed-loop navigation using single-plane fluoroscopic tracking without requiring biplane imaging or repeated contrast administration during navigation.

This approach provides an alternative to imaging strategies that directly recover 3D robot pose [24, 38]. Optical and stereo-imaging systems can provide accurate 3D localization in laboratory environments but are not applicable within deep vascular anatomy [39]. Biplane fluoroscopy can recover additional spatial information in vivo but requires specialized imaging infrastructure and introduces additional radiation, calibration, and workspace constraints. VISTA instead operates with widely available single-plane fluoroscopy by using the pre-existing vascular model to supply the spatial information absent from the 2D projection. This distinction shifts the imaging requirement from continuously reconstructing the robot and surrounding anatomy in 3D to detecting the robot within a known anatomical trajectory.

The same representation also provides a practical basis for autonomous magnetic control. Because the local vessel orientation is defined by consecutive centerline milestones, VISTA can automatically update the rotating magnetic-field axis as the robot progresses through the vessel. The milestone spacing establishes a trade-off between geometric fidelity and robotic control efficiency. Excessively frequent orientation updates increase manipulator motion, whereas larger intervals increase angular mismatch between the commanded actuation axis and local vessel orientation. By relating this mismatch to magnetic-field and torque retention, the present framework provides an actuation-based criterion for selecting milestone spacing rather than treating trajectory discretization as an arbitrary computational parameter. This integration of localization and actuation within the same anatomical representation is particularly important for untethered systems, in which imaging and magnetic control cannot be considered independently.

Autonomous control also changes the role of the human operator. Manual fluoroscopic navigation requires simultaneous interpretation of 2D imaging, inference of motion relative to 3D vascular anatomy, and manipulation of the external magnetic actuator under flow. In the present experiments, VISTA reduced navigation time, corrective actuation commands, and X-ray exposure relative to representative novice and experienced manual operators. These comparisons should be interpreted as a controlled benchmark rather than a general comparison with clinical operator performance, but they demonstrate the potential advantage of shifting the operator from continuous low-level magnetic manipulation toward supervisory control. Additional experiments under altered field orientation and rotational speed further showed that effective propulsion depends strongly on maintaining an appropriate relationship between the magnetic actuation axis and local vascular geometry, particularly in tortuous trajectories. Automated use of the digital-twin geometry provides a systematic means of maintaining this relationship that becomes increasingly difficult to reproduce through manual control alone.

Reduced fluoroscopic exposure represents an additional clinically relevant benefit of autonomous navigation. Fluoroscopy-guided cardiovascular and endovascular procedures can result in substantial radiation exposure to both patients and operators, with reported reference air kerma values ranging from hundreds to more than one thousand mGy depending on procedural complexity [40, 41]. Although these procedure-level values are not directly comparable with the task-level exposures measured in our phantom experiments, fluoroscopy time remains a modifiable contributor to radiation burden [42]. VISTA reduced X-ray exposure during navigation by up to 57% compared with manual control by shortening navigation time and reducing corrective maneuvers.

An important translational feature of VISTA is the separation of real-time robot tracking from vascular visualization. Fluoroscopy tracks the radiopaque microrobot, while the registered 3D digital twin provides the vascular topology required for navigation, eliminating the need for repeated contrast administration as long as the underlying vascular geometry remains sufficiently stable. Although contrast-enhanced rotational angiography was used to construct the digital twins in this study, VISTA is agnostic to the source of the vascular model, which could clinically be obtained from preprocedural CT or MR angiography, including non-contrast MR angiography when iodinated contrast is undesirable. The present implementation, however, represents the vasculature as a static anatomical model and therefore does not capture intraoperative changes in vessel geometry or physiology. More advanced digital twins could overcome this limitation by incorporating real-time physiological and imaging data to update the virtual representation as conditions change [43]. Such inputs could include ultrasound-derived vessel geometry and flow, cardiac or respiratory motion, and other intraoperative measurements, enabling adaptive registration and navigation while preserving the advantage of minimizing contrast use.

Integration of preprocedural digital twin construction with intraoperative fluoroscopy is also compatible with existing interventional imaging workflows. Existing angiographic platforms already support registration and fusion of CT- or MR-derived 3D anatomical datasets with live 2D fluoroscopy, and related image-fusion approaches have been used to reduce fluoroscopy time, radiation exposure, or contrast administration during endovascular procedures [44-47] [48, 49] [50, 51] [50, 52-56]. VISTA builds on this established paradigm but uses the registered anatomical model as an active component of robotic localization and control rather than solely as a visual roadmap. The same patient-specific vascular models could additionally be used for preprocedural phantom fabrication and rehearsal, allowing challenging vascular geometries, robot-arm workspace constraints, and magnetic actuation limitations to be identified before intervention [57].

A distinct limitation arises from the physical accessibility of the external magnetic actuator. Although VISTA provides the spatial information required to determine the desired magnetic actuation axis, successful navigation additionally requires the robot-mounted magnet to physically achieve the corresponding position and orientation. This constraint becomes increasingly important at human scale, where anatomical depth, vessel tortuosity, robot-arm kinematics, patient clearance, and potential interference with the fluoroscopic imaging system collectively restrict the accessible magnetic workspace [58]. Consequently, resolving depth ambiguity through VISTA does not guarantee that every desired actuation configuration is physically achievable. Future implementations should therefore incorporate actuator reachability and magnetic-field feasibility directly into navigation planning, allowing candidate trajectories and actuation states to be evaluated against both vascular geometry and the available magnetic workspace.

The current digital twin also represents vascular anatomy as static, whereas vessels undergo deformation due to cardiac pulsation, respiratory motion, vasodilation, vasoconstriction, and other physiological processes. These changes can introduce discrepancies between the preprocedural model and intraoperative anatomy. Future implementations could incorporate deformable registration or physiologically informed vascular models to update the digital twin during navigation. Similarly, the present study primarily evaluates navigation along individual vessel trajectories. Extension to branched vascular networks will require graph-based milestone representations in which multiple candidate paths can be evaluated according to the current navigation state, target anatomy, and observed robot motion. Such representations could enable autonomous branch selection and path planning while retaining the topology-constrained localization principle demonstrated here.

Extension to torturous and branched vascular networks represent another important direction for future development. The present work focuses primarily on navigation within individual vessel segments and in-vivo models. With proper modifications, the underlying framework can be extended to larger vascular trees by incorporating milestone networks in branching centerline representations within the digital twin. In such cases, pose inference could be performed across multiple candidate paths, with branch selection guided by navigation goals, shortest path finding algorithm, anatomical constraints, or real-time observations of microrobot behavior.

Further reductions in fluoroscopic exposure may be possible by moving from continuous observation toward predictive, event-driven imaging. Recent fluoroscopy-guided microrobotic systems have demonstrated that adaptive Kalman filtering can maintain trajectory estimates during transient loss of visual detection, supporting the feasibility of model-based prediction between reliable observations. Because VISTA further constrains robot motion to an ordered vascular trajectory, state estimators could predict progression between milestones while fluoroscopy is inactive. Imaging could then be triggered selectively when localization uncertainty exceeds a predefined threshold or to confirm milestone transitions, rather than continuously monitoring the robot. This would extend the current event-driven control architecture to uncertainty-triggered imaging [28], potentially further reducing radiation exposure while preserving localization reliability.

Overall, VISTA demonstrates that a patient-specific vascular digital twin can provide the spatial context missing from single-plane fluoroscopy and convert limited 2D robot observations into actionable navigation states for autonomous magnetic control. By separating real-time robot tracking from continuous visualization of the surrounding vasculature, the framework enables contrast-free navigation using widely available fluoroscopic infrastructure while avoiding the requirement for unconstrained 3D pose reconstruction. More broadly, these findings suggest that clinical translation of untethered endovascular microrobots will depend not only on improvements in robot design and magnetic actuation, but also on navigation architectures that integrate prior anatomical knowledge, real-time imaging, and autonomous control. Such integration provides a pathway toward endovascular microrobotic systems in which clinicians supervise high-level therapeutic objectives while model-based control manages the underlying navigation task.

## MATERIALS AND METHOD

### *3D angiography and phantom vessel preparation*

Reconstruction of the target vessel segment was performed using 3D rotational angiography acquired with an OEC 3D C-arm (GE HealthCare, MA, USA). During acquisition, the flat-panel detector rotated approximately 200° around the contrast-filled target vessel, acquiring 512 projection images over the rotational arc. The acquired projections were reconstructed into a 3D volumetric dataset using 3D Slicer (open source, v5.2.2). The vessel was subsequently segmented to generate a 3D vascular mesh, which served as the anatomical basis of the digital twin used for VISTA localization and navigation. Within the registered digital twin, the detected 2D EndoBot position was associated with the projected vascular centerline to estimate its corresponding 3D navigation state and local vessel orientation.

In vitro phantom vessels were designed using Fusion 360 (Autodesk, San Francisco, CA, USA) and fabricated using SYLGARD 184 Silicone Elastomer Base and Curing Agent (Dow Chemical Company, Midland, MI, USA) and sacrificial molds printed with a K1 Max 3D printer (Creality, Shenzhen, China) using acrylonitrile butadiene styrene (ABS) filament. Following the curing of SYLGARD 184, ABS filaments were dissolved with acetone. Phantom Models 2 and 3 in Table S1 were derived from the same umbilical cord used in our previous work [10] but configured into different geometries to evaluate the system under varying three-dimensional vascular conditions.

***Creation and evaluation of object detection model***

To obtain the EndoBot center point and bounding box coordinates, an object detection model was trained using the Roboflow platform (Roboflow, Inc., Des Moines, Iowa, United States) [59] with the You Only Look Once algorithm (YOLOv8, Ultralytics, MD, United States) [60]. The dataset consisted of 817 distinct x-ray and RGB camera images and was divided into training, validation, and test sets. The training set included 562 images, corresponding to 68% of the dataset; the validation set included 131 images, corresponding to 16%; and the test set included 124 images, corresponding to 15%. During training, the 562 training images were subjected to online augmentation, including scaling, flipping, mosaic augmentation, rotation, blurring, saturation adjustment, brightness adjustment, exposure variation, and grayscale conversion. Although these augmentations could be applied offline to generate an expanded static dataset of approximately 3,240 images, we instead applied them online during training. In this approach, each epoch used the original 562 training images while applying randomized augmentation parameters, generating a different augmented representation of the training set at each epoch. Therefore, over 50 training epochs, the model was exposed to 28,100 augmented image presentations, improving dataset variability without permanently increasing the stored dataset size.

***θ angle analysis and implementation***

To incorporate robot-arm motion and magnet orientation into the angular deviation analysis, a second set of offset “actuator milestones” was generated from the original vessel milestones using predefined positional and rotational margins. These actuator milestones allowed the external magnet to follow the vessel trajectory while preserving the desired magnet-EndoBot coupling and avoiding interference with the experimental setup. Simulated robot motion through these milestones was then used to quantify angular deviation between the propagated magnet orientation and the local vessel milestone orientation.

***Robot arm external Control and ROS2 data transmission***

Physical magnetic actuation was performed using an LBR Med 7 R800 robot arm (KUKA AG, Augsburg, Germany) controlled through the Fast Robot Interface (FRI). Unlike conventional Sunrise-based KUKA control, the FRI connection enabled real-time external motion commands from a C++ application through the KUKA Optional Network Interface (KONI) port. This allowed the C-arm fluoroscopy system, object detection pipeline, Unity virtual environment, and KUKA robot arm to operate as a unified autonomous navigation platform.

A ROS2-based communication network was used to exchange data between the object detection pipeline, Unity simulation, and robot control system. The detection pipeline published environmental and EndoBot tracking information, while Unity converted these inputs into the final estimated position and orientation and transmitted them through a topic names "posrot". A C++ subscriber received the latest data over "posrot" topic and incorporated it into the FRI control loop, allowing the robot arm to update the external magnet pose in real time.

Target robot poses were calculated relative to the initial robot configuration and updated using the latest Unity-derived position and orientation estimates. Robot motion was executed using linear interpolation for position and spherical linear interpolation for orientation, allowing smooth transitions between target poses. With each new estimation, system directly updates the target position and starts robot arm movement towards to the next target pose, ensuring that both positional and rotational errors remained below predefined thresholds (Supplementary Text 5).

***Unity setup and experimental fixes***

To improve visualization and facilitate robot-arm positioning, the virtual magnet in Unity was assigned to a separate milestone set derived from the original vessel centerline. The position and orientation of this magnet-specific milestone set can be offset relative to the original EndoBot milestone set according to the dimensions and placement of the physical magnet. By adjusting these translational and rotational offsets, the robot-arm trajectory can be configured to maintain an appropriate magnet position and preserve the desired magnetic-field orientation relative to the EndoBot and vessel geometry. All associated parameters can be modified through the Unity control panel, allowing the virtual setup to be adapted to different phantom geometries and experimental configurations.

***Statistical Analysis***

Statistical analysis in **Fig. 6** was performed using three independent standard-condition trials. For each trial, total navigation time, X-ray exposure, and movement count were calculated by summing measurements from the forward and backward navigations. Total navigation metrics were used for performance comparison. Data are reported as mean ± standard deviation. Pairwise comparisons between groups were performed using one-sided, unpaired Student's t-tests assuming equal variances. Statistical significance was defined as $p < 0.05$.

**Acknowledgments**

Funding: H.C. acknowledges financial support for this study from Career Development Award in Cardiovascular Disease Research Honoring Dr. Earl W. Wood, administered by the Mayo Clinic Center for Clinical and Translational Science; the State of Arizona, Arizona Biomedical Research Centre New Investigator Award (award number RFGA2024-022-002); and the National Heart, Lung, and Blood Institute of the National Institutes of Health under Award Number R01HL179093).

**Author Contributions**

**Husnu Halid Alabay:** Investigation, methodology, data curation, formal analysis, visualization, writing – original draft, writing – review & editing.

**Tuan Anh Le:** Investigation, data curation, formal analysis, writing – review & editing.

**Ping Wang:** Investigation, validation, writing – review & editing.

**Hakan Ceylan:** Conceptualization, supervision, project administration, funding acquisition, resources, visualization, writing – original draft, writing – review & editing.

**Conflict of Interest Statement**

H.H.A., T.-A.L., and H.C. filed a patent for the method presented herein.

**Data and Materials Availability**

All data needed to evaluate the conclusions in the paper are presented in the paper. Additional data is available from the corresponding author upon reasonable request.

**Declaration of generative AI and AI-assisted technologies**

During the preparation of this work the authors used Open AI, ChatGPT 5.2-5.6 to improve language clarity and organization. The authors verified and edited all AI-assisted content, and all ideas, analyses, and final conclusions are entirely those of the authors. AI tools were not used for data generation, analysis, interpretation, or substantive scientific writing.

## References

[1] M. Sitti, H. Ceylan, W. Hu *et al.*, "Biomedical applications of untethered mobile milli/microrobots," *Proceedings of the IEEE,* vol. 103, no. 2, pp. 205-224, 2015.

[2] H. Ceylan, J. Giltinan, K. Kozielski *et al.*, "Mobile microrobots for bioengineering applications," *Lab on a Chip,* vol. 17, no. 10, pp. 1705-1724, 2017.

[3] J. Li, B. Esteban-Fernández de Ávila, W. Gao *et al.*, "Micro/nanorobots for biomedicine: Delivery, surgery, sensing, and detoxification," *Science robotics,* vol. 2, no. 4, pp. eaam6431, 2017.

[4] F. Soto, and R. Chrostowski, "Frontiers of medical micro/nanorobotics: in vivo applications and commercialization perspectives toward clinical uses," *Frontiers in bioengineering and biotechnology,* vol. 6, pp. 170, 2018.

[5] A. Aziz, S. Pane, V. Iacovacci *et al.*, "Medical imaging of microrobots: Toward in vivo applications," *ACS nano,* vol. 14, no. 9, pp. 10865-10893, 2020.

[6] P. E. Dupont, B. J. Nelson, M. Goldfarb *et al.*, "A decade retrospective of medical robotics research from 2010 to 2020," *Science robotics,* vol. 6, no. 60, pp. eabi8017, 2021.

[7] B. Wang, K. Kostarelos, B. J. Nelson *et al.*, "Trends in micro-/nanorobotics: materials development, actuation, localization, and system integration for biomedical applications," *Advanced Materials,* vol. 33, no. 4, pp. 2002047, 2021.

[8] X. Ju, C. Chen, C. M. Oral *et al.*, "Technology Roadmap of Micro/Nanorobots," *ACS Nano,* vol. 19, no. 27, pp. 24174-24334, Jul 15, 2025.

[9] H. Ceylan, E. Sinibaldi, S. Misra *et al.*, "How microrobots should be translated: A clinical and value-centered readiness framework," *Bioengineering & Translational Medicine*, pp. e70112, 2026.

[10] T. A. Le, H. H. Alabay, P. G. Singh *et al.*, "Physiologically adaptive soft Millirobot for atraumatic endovascular therapy," *Advanced functional materials,* vol. 36, no. 35, pp. e31400, 2026.

[11] W. Hu, G. Z. Lum, M. Mastrangeli *et al.*, "Small-scale soft-bodied robot with multimodal locomotion," *Nature,* vol. 554, no. 7690, pp. 81-85, Feb 1, 2018.

[12] X. Hu, I. C. Yasa, Z. Ren *et al.*, "Magnetic soft micromachines made of linked microactuator networks," *Sci Adv,* vol. 7, no. 23, pp. eabe8436, Jun, 2021.

[13] F. C. Landers, L. Hertle, V. Pustovalov *et al.*, "Clinically ready magnetic microrobots for targeted therapies," *Science,* vol. 390, no. 6774, pp. 710-715, Nov 13, 2025.

[14] S. A. Abbasi, A. Ahmed, S. Noh *et al.*, "Autonomous 3D positional control of a magnetic microrobot using reinforcement learning," *Nature Machine Intelligence,* vol. 6, no. 1, pp. 92-105, Jan, 2024.

[15] S. Çimen, A. Gooya, M. Grass *et al.*, "Reconstruction of coronary arteries from X-ray angiography: A review," *Medical image analysis,* vol. 32, pp. 46-68, 2016.

[16] N. E. Shalom, G. X. Gong, and M. Auster, "Fluoroscopy: An essential diagnostic modality in the age of high-resolution cross-sectional imaging," *World J Radiol,* vol. 12, no. 10, pp. 213-230, Oct 28, 2020.

[17] M. E. Abdelaziz, L. Tian, M. Hamady *et al.*, “X-ray to MR: The progress of flexible instruments for endovascular navigation,” *Progress in Biomedical Engineering,* vol. 3, no. 3, pp. 032004, 2021.

[18] B. Hennessey, H. Danenberg, F. De Vroey *et al.*, “Dynamic Coronary Roadmap versus standard angiography for percutaneous coronary intervention: the randomised, multicentre DCR4Contrast trial,” *EuroIntervention,* vol. 20, no. 3, pp. e198, 2024.

[19] M. S. Davenport, M. A. Perazella, J. Yee *et al.*, “Use of intravenous iodinated contrast media in patients with kidney disease: consensus statements from the American College of Radiology and the National Kidney Foundation,” *Radiology,* vol. 294, no. 3, pp. 660-668, 2020.

[20] R. Mehta, D. Sorbo, F. Ronco *et al.*, “Key Considerations regarding the Renal Risks of Iodinated Contrast Media: The Nephrologist's Role,” *Cardiorenal Med,* vol. 13, no. 1, pp. 324-331, 2023.

[21] D. Bertsche, V. Rasche, W. Rottbauer *et al.*, “3D localization from 2D X-ray projection,” *International Journal of Computer Assisted Radiology and Surgery,* vol. 17, no. 9, pp. 1553-1558, 2022.

[22] A. Ramadani, M. Bui, T. Wendler *et al.*, “A survey of catheter tracking concepts and methodologies,” *Medical image analysis,* vol. 82, pp. 102584, 2022.

[23] W. Peng, W. Wu, J. Zhang *et al.*, “An automatic framework for estimating the pose of the catheter distal section using a coarse-to-fine network,” *Computer Methods and Programs in Biomedicine,* vol. 225, pp. 107036, 2022.

[24] P. B. Nguyen, B. Kang, D. Bappy *et al.*, “Real-time microrobot posture recognition via biplane X-ray imaging system for external electromagnetic actuation,” *International journal of computer assisted radiology and surgery,* vol. 13, no. 11, pp. 1843-1852, 2018.

[25] J.-H. Kim, K.-G. Kim, H.-S. Lee *et al.*, "Development of bi-plane X-ray imaging system for real-time position recognition of intravascular therapeutic microrobot using EMA system." pp. 1-2.

[26] J. Hwang, S. Jeon, B. Kim *et al.*, “An Electromagnetically Controllable Microrobotic Interventional System for Targeted, Real-Time Cardiovascular Intervention,” *Advanced healthcare materials,* vol. 11, no. 11, pp. 2102529, 2022.

[27] P. B. Nguyen, J.-O. Park, S. Park *et al.*, "Medical micro-robot navigation using image processing-blood vessel extraction and X-ray calibration." pp. 365-370.

[28] C. Wang, W. Kang, M. Sun *et al.*, “Synthetic X-ray-driven tracking and control of miniature medical devices,” *Nature Machine Intelligence,* vol. 8, no. 2, pp. 276-291, 2026.

[29] K. A. Guru, S. B. Shafiei, A. Khan *et al.*, “Understanding cognitive performance during robot-assisted surgery,” *Urology,* vol. 86, no. 4, pp. 751-757, 2015.

[30] E. Lau, N. A. Alkhamesi, and C. M. Schlachta, “Impact of robotic assistance on mental workload and cognitive performance of surgical trainees performing a complex minimally invasive suturing task,” *Surgical endoscopy,* vol. 34, no. 6, pp. 2551-2559, 2020.

[31] X. Tang, Y. Li, X. Liu *et al.*, “Vision-based automated control of magnetic microrobots,” *Micromachines,* vol. 13, no. 2, pp. 337, 2022.

[32] L.-J. W. Ligtenberg, M. C. De Boer, I. Mulder *et al.*, "X-Ray-Guided Magnetic Fields for Wireless Control of Untethered Magnetic Robots in Cerebral Vascular Phantoms." pp. 4624-4629.

[33] H. H. Alabay, T.-A. Le, and H. Ceylan, "X-ray fluoroscopy guided localization and steering of miniature robots using virtual reality enhancement," *Frontiers in Robotics and AI,* vol. 11, pp. 1495445, 2024.

[34] T. A. Le, H. H. Alabay, P. G. Singh *et al.*, "Physiologically Adaptive Soft Millirobot for Atraumatic Endovascular Therapy," *Advanced Functional Materials*, pp. e31400, 2026.

[35] P. K. Saha, G. Borgefors, and G. S. di Baja, "A survey on skeletonization algorithms and their applications," *Pattern recognition letters,* vol. 76, pp. 3-12, 2016.

[36] KUKA AG. "LBR Med: Lightweight cobot for medical applications," August 20, 2026; https://www.kuka.com/en-ca/industries/robots-in-medicine/kuka-solutions-for-medical-robots/lbr-med-medical-robot.

[37] P. Wang, T.-A. Le, P. Singh *et al.*, "Non-survival rat endovascular testbed for early-stage evaluation of untethered magnetic microrobots," *MethodsX*, pp. 103906, 2026.

[38] J. P. Bae, S. Yoon, M. Vania *et al.*, "Three dimensional microrobot tracking using learning-based system," *International Journal of Control, Automation and Systems,* vol. 18, no. 1, pp. 21-28, 2020.

[39] S. Mohanty, A. Hong, C. Alcantara *et al.*, "Stereo holographic diffraction based tracking of microrobots," *IEEE Robotics and Automation Letters,* vol. 3, no. 1, pp. 567-572, 2017.

[40] R. Sánchez, E. Vañó, J. M. F. Soto *et al.*, "Updating national diagnostic reference levels for interventional cardiology and methodological aspects," *Physica Medica,* vol. 70, pp. 169-175, 2020.

[41] V. Kataria, I. Yaduvanshi, G. Singal *et al.*, "Establishing a diagnostic reference level of radiation dose in coronary angiography and intervention: A prospective evaluation," *Indian Heart Journal,* vol. 73, no. 6, pp. 725-728, 2021.

[42] R. Fazel, J. Curtis, Y. Wang *et al.*, "Determinants of fluoroscopy time for invasive coronary angiography and percutaneous coronary intervention: Insights from the NCDR®," *Catheterization and Cardiovascular Interventions,* vol. 82, no. 7, pp. 1091-1105, 2013.

[43] W. Trevena, X. Zhong, A. Lal *et al.*, "Model-driven engineering for digital twins: a graph model-based patient simulation application," *Frontiers in Physiology,* vol. 15, pp. 1424931, 2024.

[44] Siemens Healthineers. "Syngo 3D Roadmap and syngo Toolbox," July 9, 2026; https://www.siemens-healthineers.com/en-us/angio/options-and-upgrades/clinical-software-applications/syngo-ipilot.

[45] Siemens Healthineers. "Syngo Fusion Package," July 9, 2026; https://www.siemens-healthineers.com/angio/options-and-upgrades/clinical-software-applications/syngo-fusion-package.

[46] Siemens Healthineers. "Syngo iFlow," July 9, 2026; https://www.siemens-healthineers.com/angio/options-and-upgrades/clinical-software-applications/syngo-iflow.

[47] M. Tang, F. Zeng, X. Chang *et al.*, “Feasibility study of Syngo iFlow in predicting hemodynamic improvement post-endovascular procedure in peripheral artery disease,” *BMC Cardiovascular Disorders,* vol. 24, no. 1, pp. 99, 2024.

[48] Philips. "XperGuide: Live 3D needle guidance," July 9, 2026; https://www.usa.philips.com/healthcare/product/HCOPT06/xperguide-live-3d-needle-guidance.

[49] D. Fior, F. Vacirca, D. Leni *et al.*, “Virtual guidance of percutaneous transthoracic needle biopsy with C-arm cone-beam CT: diagnostic accuracy, risk factors and effective radiation dose,” *CardioVascular and Interventional Radiology,* vol. 42, no. 5, pp. 712-719, 2019.

[50] V. Tacher, M. Lin, P. Desgranges *et al.*, “Image guidance for endovascular repair of complex aortic aneurysms: comparison of two-dimensional and three-dimensional angiography and image fusion,” *Journal of Vascular and Interventional Radiology,* vol. 24, no. 11, pp. 1698-1706, 2013.

[51] D.-K. Jang, D. A. Stidd, S. Schafer *et al.*, “Monoplane 3D overlay roadmap versus conventional biplane 2D roadmap technique for neurointervenional procedures,” *Neurointervention,* vol. 11, no. 2, pp. 105-113, 2016.

[52] D. W. Jones, L. Stangenberg, N. J. Swerdlow *et al.*, “Image fusion and 3-dimensional roadmapping in endovascular surgery,” *Annals of Vascular Surgery,* vol. 52, pp. 302-311, 2018.

[53] C. J. Schulz, D. Böckler, J. Krisam *et al.*, “Two-dimensional-three-dimensional registration for fusion imaging is noninferior to three-dimensional-three-dimensional registration in infrarenal endovascular aneurysm repair,” *Journal of Vascular Surgery,* vol. 70, no. 6, pp. 2005-2013, 2019.

[54] M. Sieren, C. Schareck, M. Kaschwich *et al.*, “Accuracy of registration techniques and vascular imaging modalities in fusion imaging for aortic endovascular interventions: a phantom study,” *CVIR endovascular,* vol. 4, no. 1, pp. 51, 2021.

[55] S. P. Smorenburg, R. J. Lely, I. Smit-Ockeloen *et al.*, “Automated image fusion during endovascular aneurysm repair: a feasibility and accuracy study,” *International Journal of Computer Assisted Radiology and Surgery,* vol. 18, no. 8, pp. 1533-1541, 2023.

[56] B. J. Mittmann, A. Seitel, G. Echner *et al.*, “Reattachable fiducial skin marker for automatic multimodality registration,” *International journal of computer assisted radiology and surgery,* vol. 17, no. 11, pp. 2141-2150, 2022.

[57] J. Coles-Black, D. Bolton, and J. Chuen, "Accessing 3D printed vascular phantoms for procedural simulation. Front Surg. 2020; 7: 626212."

[58] T. A. Le, H. H. Alabay, A. Sivarasa *et al.*, “Human-Scale Rotating Magnetic Field Delivery With Permanent Magnets Under Anatomic and Robotic Constraints,” *Advanced Robotics Research*, pp. e70154, 2026.

[59] B. Dwyer, J. Nelson, T. Hansen *et al.*, “Roboflow (Version 1.0)[Software]. 2024,” *Computer vision platform*, 2024.

[60] G. Jocher, A. Chaurasia, and J. Qiu, "YOLO by Ultralytics," 2023.

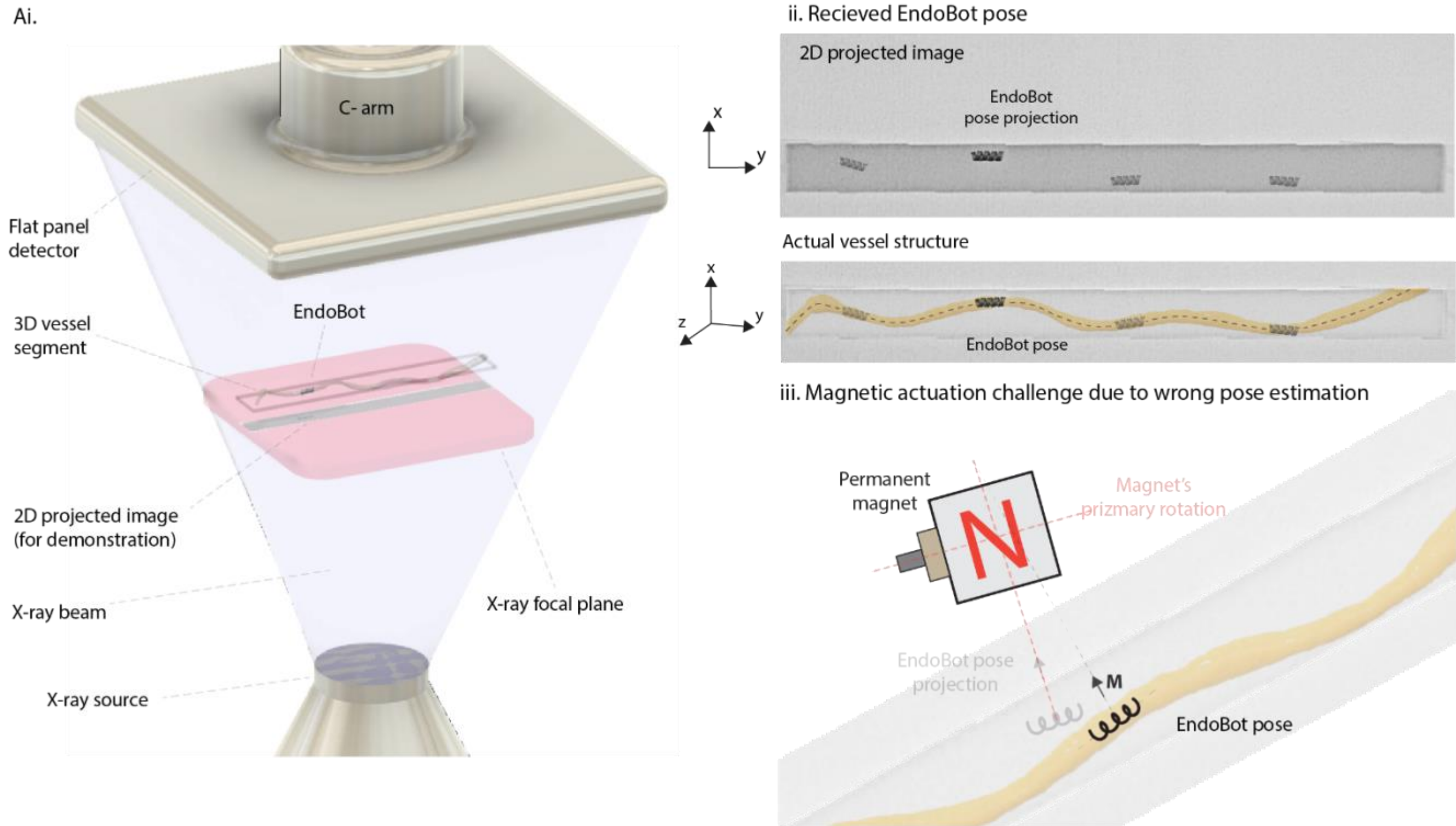
Ai.
C- arm
Flat panel detector
3D vessel segment
EndoBot
2D projected image (for demonstration)
X-ray beam
X-ray focal plane
X-ray source
ii. Recieved EndoBot pose
2D projected image
EndoBot pose projection
Actual vessel structure
EndoBot pose
iii. Magnetic actuation challenge due to wrong pose estimation
Permanent magnet
Magnet's prizmary rotation
N
EndoBot pose projection
M
EndoBot pose

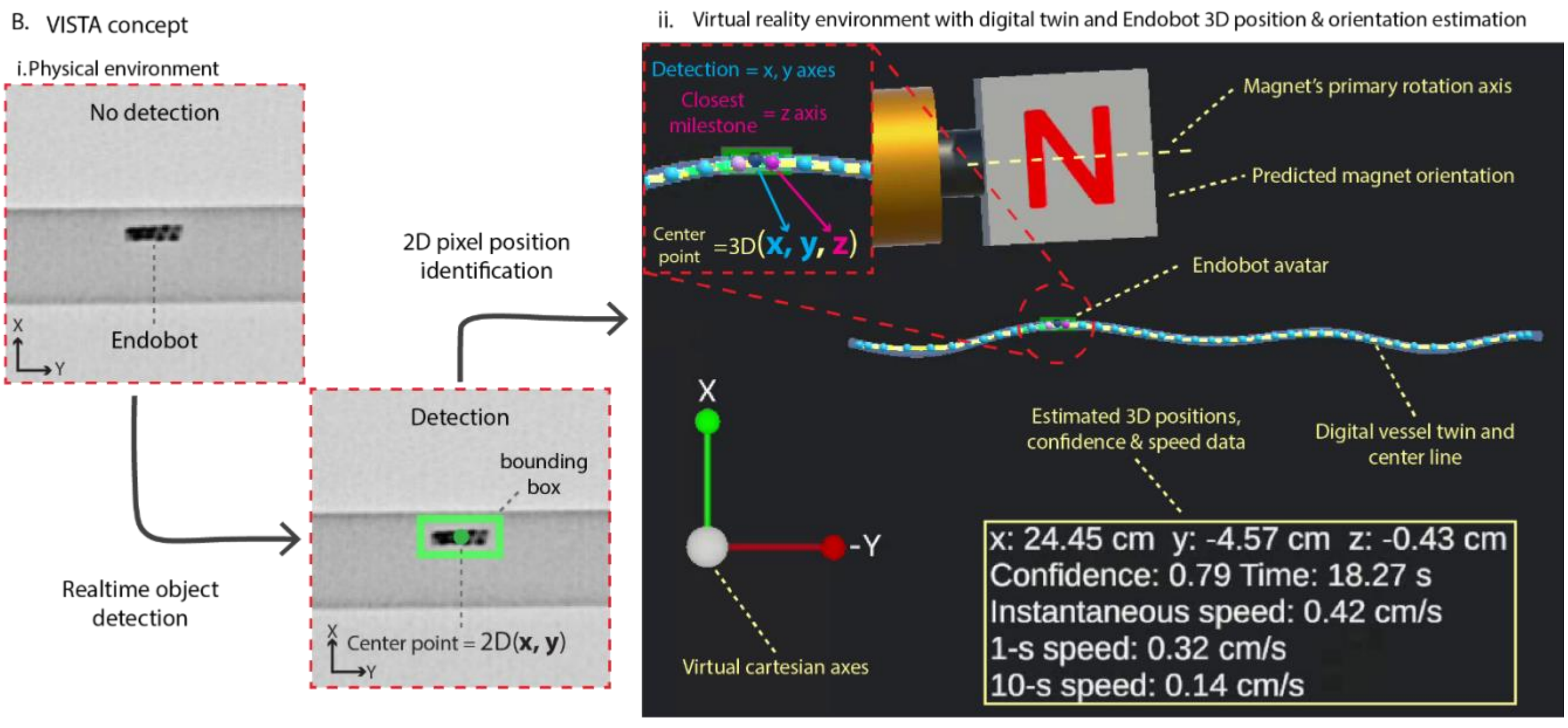
B. VISTA concept
i. Physical environment
No detection
Endobot
Realtime object detection
Detection
bounding box
Center point = 2D(x, y)
2D pixel position identification
ii. Virtual reality environment with digital twin and Endobot 3D position & orientation estimation
Detection = x, y axes
Closest milestone = z axis
Center point =3D(x, y, z)
Magnet's primary rotation axis
Predicted magnet orientation
Endobot avatar
Estimated 3D positions, confidence & speed data
Digital vessel twin and center line
Virtual cartesian axes
x: 24.45 cm y: -4.57 cm z: -0.43 cm
Confidence: 0.79 Time: 18.27 s
Instantaneous speed: 0.42 cm/s
1-s speed: 0.32 cm/s
10-s speed: 0.14 cm/s

**Figure 1 - Virtual localization framework for fluoroscopy-guided navigation of an untethered magnetic microrobot.** (A) Navigation under clinical C-arm guidance. (i) A representative single-plane fluoroscopic projection is shown in which EndoBot is visible, whereas depth information is unavailable from the projection alone. (ii) The registered vessel centerline provides the geometric context for virtual localization. (iii) Magnetic field placement creates unique challenges where angular differences between magnet and EndoBot can affect our forward displacement. (B) Overview of the VISTA framework. Fluoroscopic frames are processed using an object-detection pipeline to identify EndoBot and determine its image-plane centroid. The detected centroid is mapped onto a registered Unity-based digital twin, where the vessel centerline and predefined navigation milestones determine the current navigation state. The framework additionally reports localization confidence and navigation metrics, including instantaneous and time-averaged velocity, and communicates the navigation state with the robot arm to update the magnetic actuation.

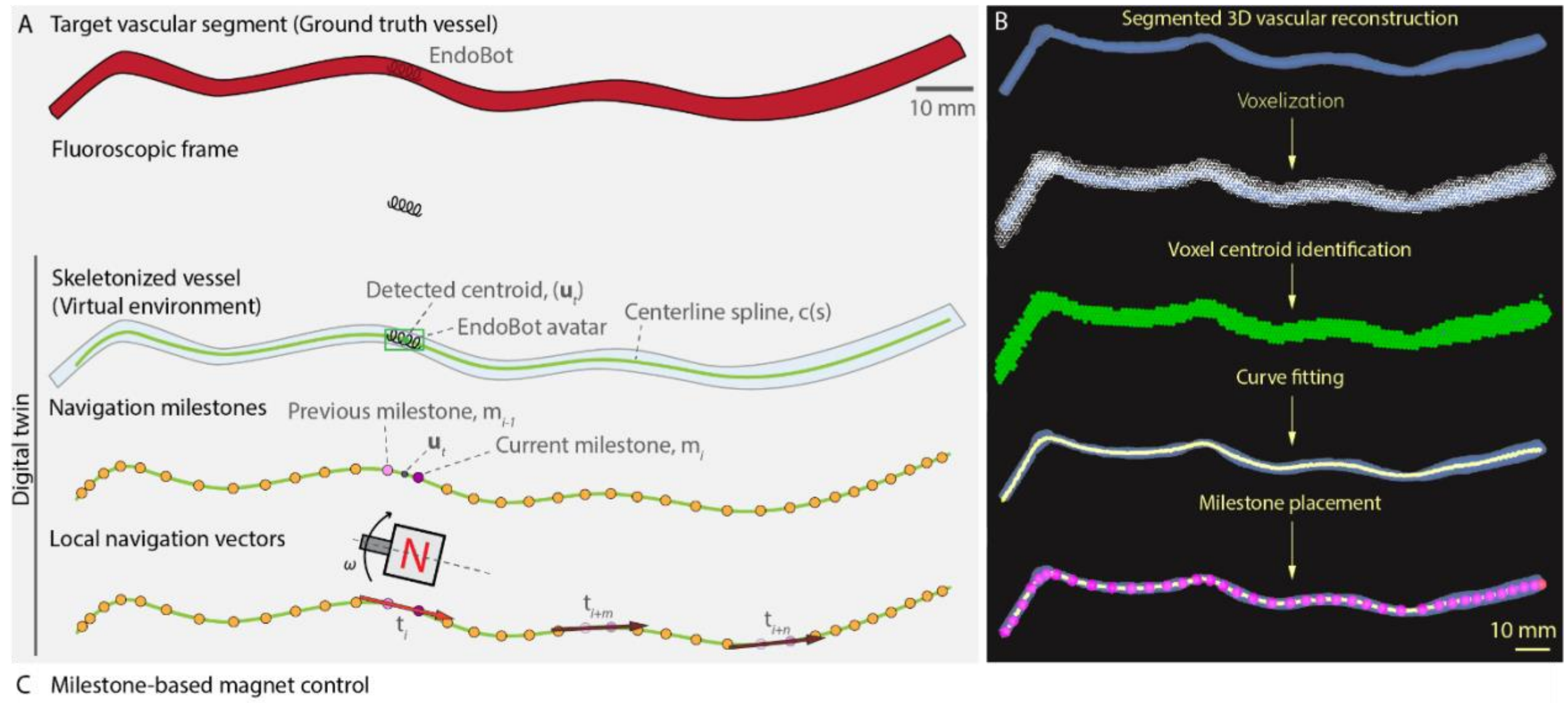

A Target vascular segment (Ground truth vessel)
EndoBot
10 mm
Fluoroscopic frame
Skeletonized vessel (Virtual environment)
Detected centroid, (u_t)
EndoBot avatar
Centerline spline, c(s)
Digital twin
Navigation milestones
Previous milestone, m_{i-1}
Current milestone, m_i
Local navigation vectors
B Segmented 3D vascular reconstruction
Voxelization
Voxel centroid identification
Curve fitting
Milestone placement
10 mm
C Milestone-based magnet control
Magnet pose held fixed during milestone interval
Next magnet pose
Effective magnetic torque, τ
Milestone interval
Previous milestone
Local navigation vector
Inactive milestone
EndoBot detected centroid
Current (assigned) milestone
Milestone re-assignment point
Inactive milestone
Next local navigation vector

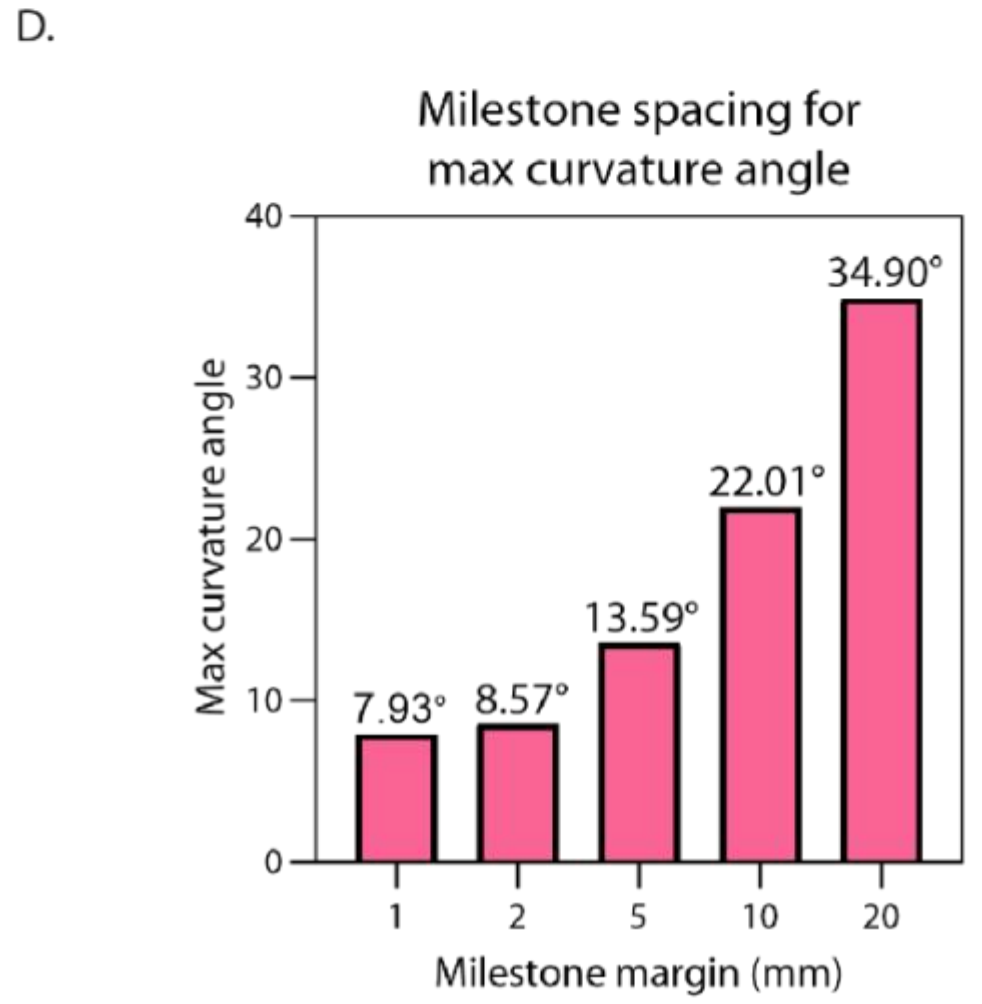

D.
Milestone spacing for max curvature angle
Max curvature angle
Milestone margin (mm)
7.93°
8.57°
13.59°
22.01°
34.90°

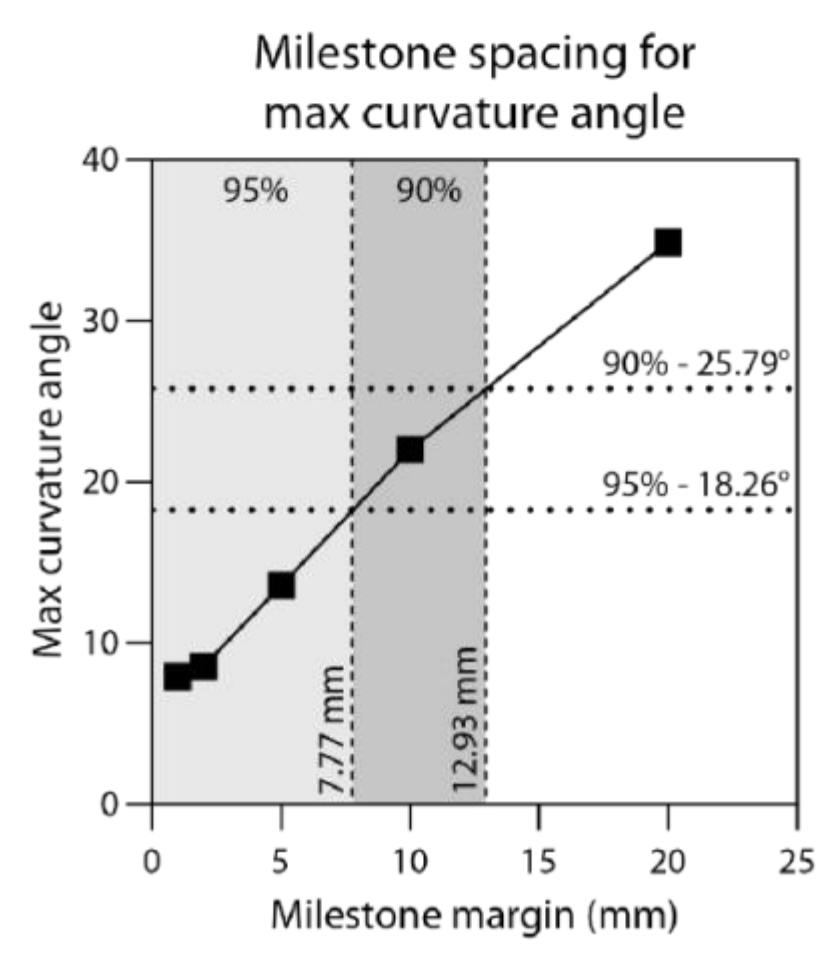

E.
Milestone spacing for max curvature angle
95%
90%
90% - 25.79°
95% - 18.26°
7.77 mm
12.93 mm
Max curvature angle
Milestone margin (mm)

**Figure 2 - Digital-twin construction and milestone-based spatial localization.**

(A) Milestone-based estimation of the EndoBot pose and local vessel direction. The detected EndoBot position was associated with the nearest centerline milestone. The relative positions of consecutive milestones defined the local vessel trajectory and were used to determine the desired orientation of the external magnetic field. (B) Generation of the vessel centerline and navigation milestones. The reconstructed vessel lumen was voxelized, and the centers of the resulting voxels were used to generate a smooth vascular centerline. Uniformly spaced milestones were then distributed along the centerline to provide discrete spatial references for localization and navigation. (C) EndoBot–milestone association and milestone initialization. The detected EndoBot position was mapped onto the generated centerline. The distance between the EndoBot position and each milestone was calculated to identify the closest milestone and initialize the corresponding milestone index. During navigation, the active milestone index was updated according to EndoBot movement. When the EndoBot crossed the midpoint between two consecutive milestones, the active index transitioned to the next milestone. This transition also updated the desired magnetic-field orientation according to the change in the local navigation vector, thereby maintaining forward propulsion along the vessel trajectory. (D) Milestone-spacing analysis. Maximum angular deviations between consecutive navigation vectors were evaluated across the tested vascular phantoms for different milestone spacings. (E) Determination of the maximum allowable milestone spacing based on magnetic-torque retention. Torque-retention thresholds were applied to the relationship between milestone spacing and maximum angular deviation. The 95% torque-retention threshold defined a maximum allowable milestone spacing of 7.77 mm, whereas the 90% torque-retention threshold defined a maximum allowable milestone spacing of 12.93 mm.

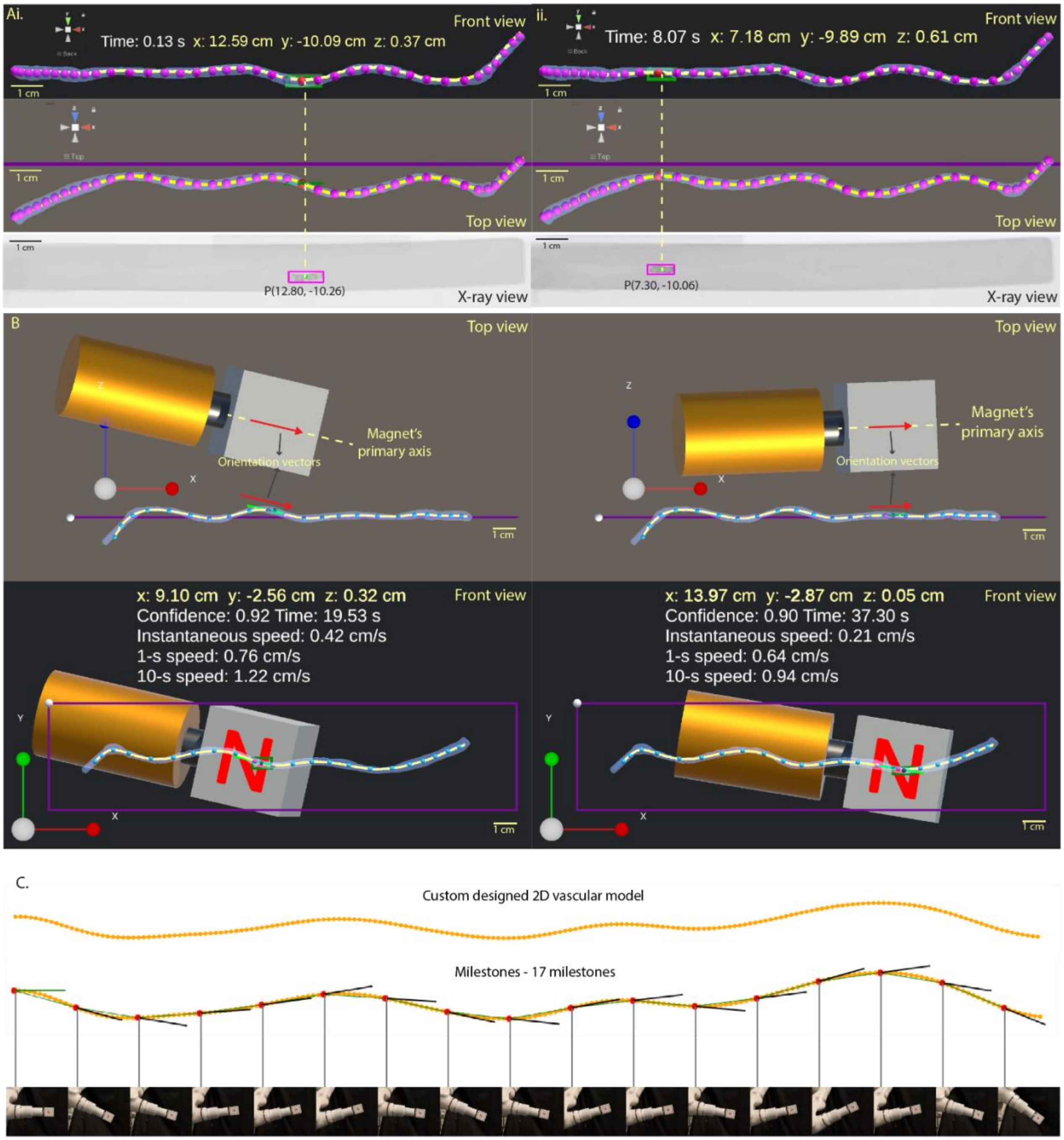
Ai.
Front view
Time: 0.13 s x: 12.59 cm y: -10.09 cm z: 0.37 cm
1 cm
Top view
X-ray view
P(12.80, -10.26)
ii.
Front view
Time: 8.07 s x: 7.18 cm y: -9.89 cm z: 0.61 cm
Top view
X-ray view
P(7.30, -10.06)
B
Top view
Magnet's primary axis
Orientation vectors
Front view
x: 9.10 cm y: -2.56 cm z: 0.32 cm
Confidence: 0.92 Time: 19.53 s
Instantaneous speed: 0.42 cm/s
1-s speed: 0.76 cm/s
10-s speed: 1.22 cm/s
x: 13.97 cm y: -2.87 cm z: 0.05 cm
Confidence: 0.90 Time: 37.30 s
Instantaneous speed: 0.21 cm/s
1-s speed: 0.64 cm/s
10-s speed: 0.94 cm/s
N
C.
Custom designed 2D vascular model
Milestones - 17 milestones

**Figure 3 - Validation of VISTA localization, orientation estimation, and robotic execution using prerecorded fluoroscopy videos.** (A) **Milestone-based localization from single plane fluoroscopy** (A) Validation of milestone-based 3D localization from single-plane fluoroscopy. Two representative EndoBot states are shown at different locations along the vessel path, corresponding to an initial position and a later position. For each state, the estimated EndoBot location is displayed in the Unity front view, Unity top view, and the corresponding X-ray image. Agreement among these views illustrates how single-plane fluoroscopic detections are mapped to the digital twin to estimate the 3D position of the EndoBot. (B) Validation of orientation estimation and magnetic-field reference generation. Using the same representative EndoBot states, the corresponding magnet orientation and EndoBot orientation are shown in Unity front and top views. These examples illustrate how the localized EndoBot state within the digital twin is used to estimate the 3D vessel direction and determine the corresponding reference orientation for magnetic actuation. (C) Validation of robotic command execution along a predefined sinusoidal trajectory. A custom sinusoidal phantom generated from 200 sampled points was discretized into 17 milestone points. The resulting centerline is shown together with the orientation states assigned to the robotic arm at each milestone. The lower panel shows representative milestone-specific images, demonstrating correspondence between the estimated local orientation and the realized orientation of the robot arm at each commanded position.

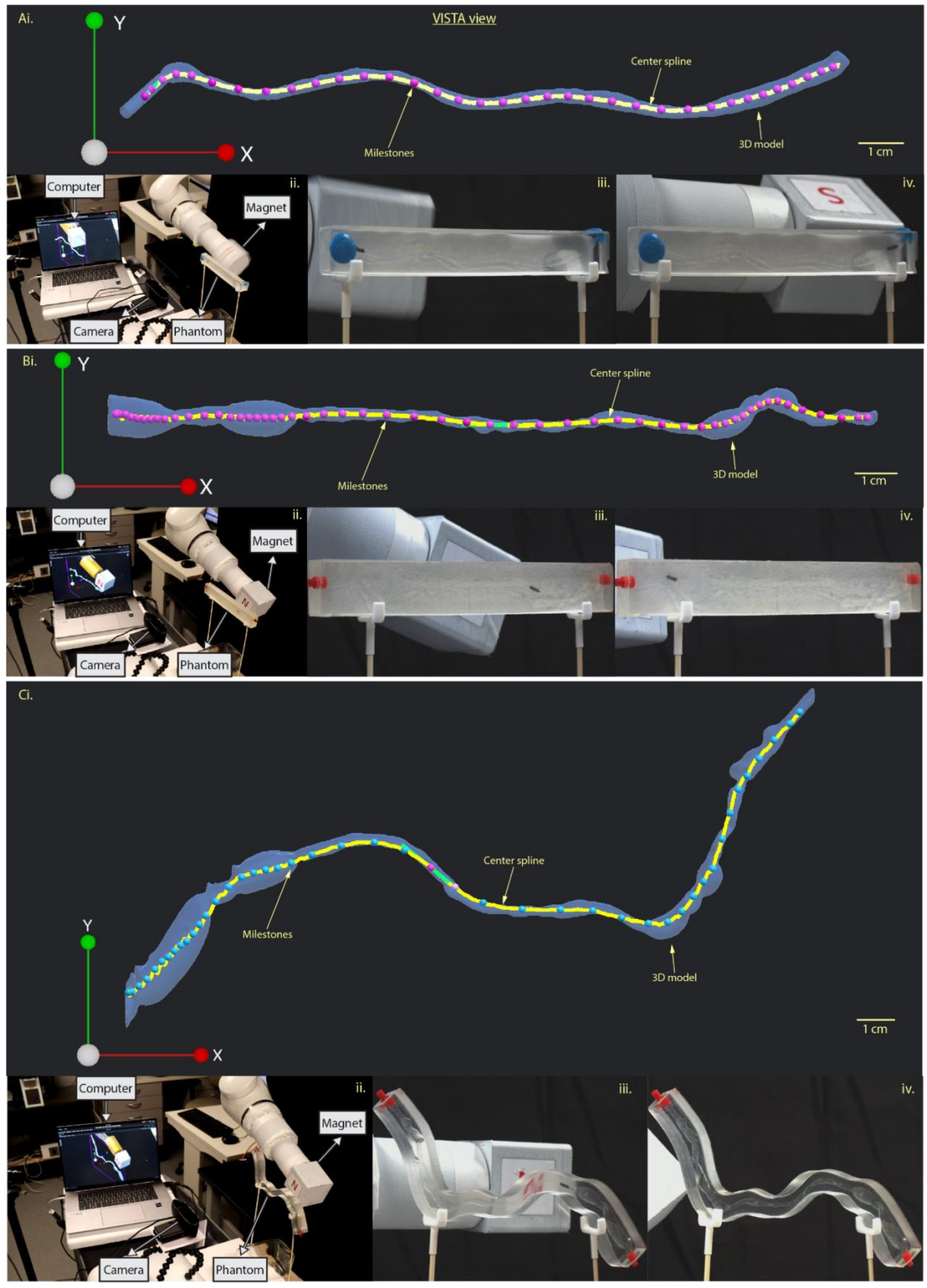
Ai.
VISTA view
Y
X
Center spline
Milestones
3D model
1 cm
Computer
Magnet
Camera
Phantom
ii.
iii.
iv.
Bi.
Y
X
Center spline
Milestones
3D model
1 cm
Computer
Magnet
Camera
Phantom
ii.
iii.
iv.
Ci.
Y
X
Center spline
Milestones
3D model
1 cm
Computer
Magnet
Camera
Phantom
ii.
iii.
iv.

**Figure 4 - VISTA-enabled autonomous navigation across umbilical vein–derived vascular phantom geometries.** (A–C) Autonomous EndoBot navigation was evaluated in three vascular phantoms with distinct geometries retrieved by umbilical cords. (A-i, B-i, C-i) Vessel-specific digital twins showing the reconstructed centerlines and assigned navigation milestones. The number and spacing of milestones were adjusted according to the geometry of each vessel to provide sufficient coverage of the navigation path. (A-ii, B-ii, C-ii) Corresponding experimental configurations showing the vascular phantom, external permanent magnet, RGB camera, and computer-based control system. (A-iii–iv, B-iii–iv, C-iii–iv) Representative fluoroscopic images showing the initial and final EndoBot positions for each phantom. In all three geometries, navigation from the inlet to the outlet was completed under fully autonomous closed-loop VISTA control.

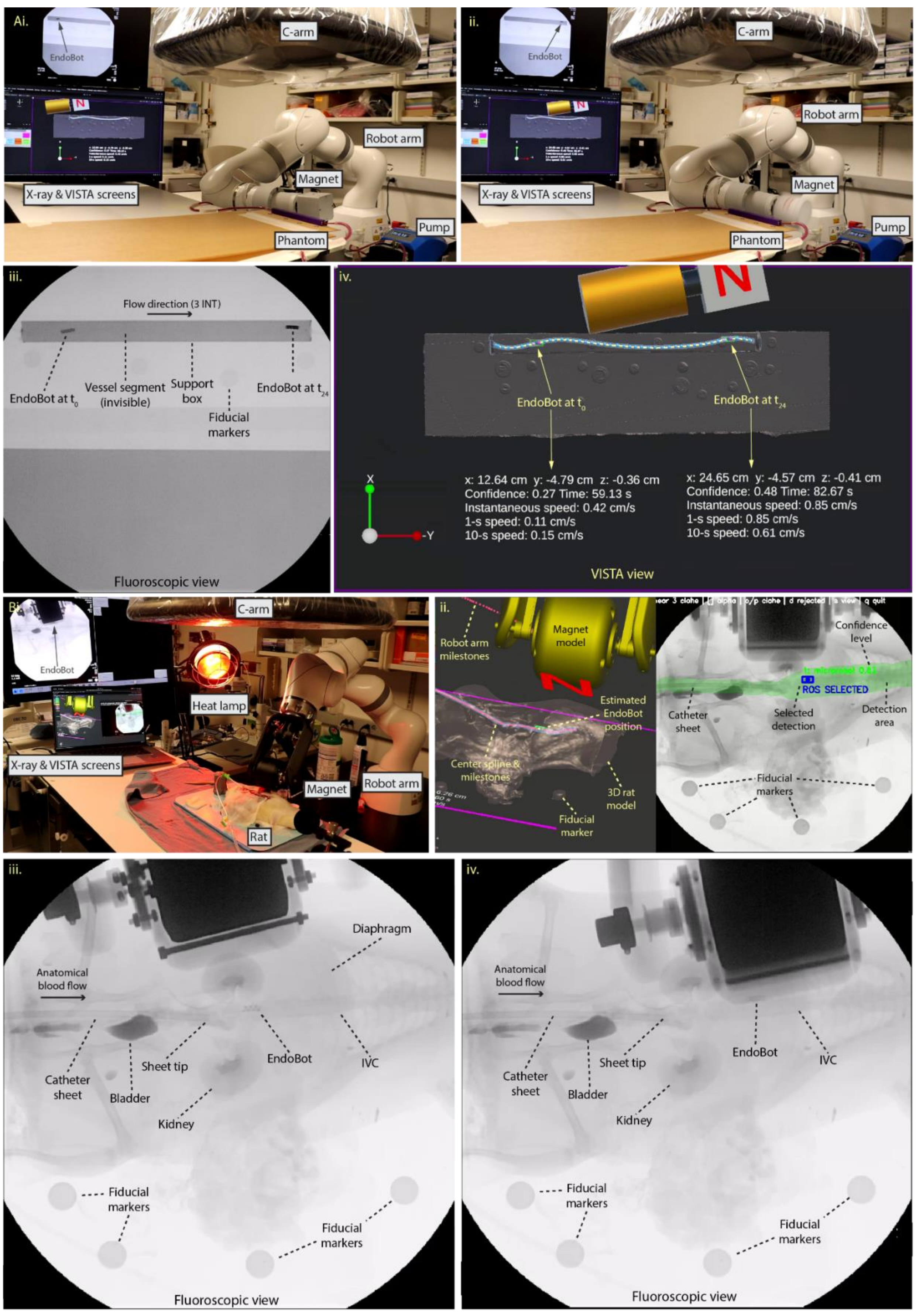
Ai.
EndoBot
C-arm
Robot arm
Magnet
X-ray & VISTA screens
Pump
Phantom
ii.
EndoBot
C-arm
Robot arm
Magnet
X-ray & VISTA screens
Pump
Phantom
iii.
Flow direction (3 INT)
EndoBot at t0
Vessel segment (invisible)
Support box
Fiducial markers
EndoBot at t24
Fluoroscopic view
iv.
EndoBot at t0
EndoBot at t24
x: 12.64 cm y: -4.79 cm z: -0.36 cm
Confidence: 0.27 Time: 59.13 s
Instantaneous speed: 0.42 cm/s
1-s speed: 0.11 cm/s
10-s speed: 0.15 cm/s
x: 24.65 cm y: -4.57 cm z: -0.41 cm
Confidence: 0.48 Time: 82.67 s
Instantaneous speed: 0.85 cm/s
1-s speed: 0.85 cm/s
10-s speed: 0.61 cm/s
X
-Y
VISTA view
Bi.
EndoBot
C-arm
Heat lamp
X-ray & VISTA screens
Magnet
Robot arm
Rat
ii.
Robot arm milestones
Magnet model
Estimated EndoBot position
Center spline & milestones
3D rat model
Fiducial marker
Confidence level
ROS SELECTED
Catheter sheet
Selected detection
Detection area
Fiducial markers
iii.
Diaphragm
Anatomical blood flow
Catheter sheet
Bladder
Sheet tip
EndoBot
IVC
Kidney
Fiducial markers
Fiducial markers
Fluoroscopic view
iv.
Anatomical blood flow
Catheter sheet
Bladder
Sheet tip
EndoBot
IVC
Kidney
Fiducial markers
Fiducial markers
Fluoroscopic view

**Figure 5 - Autonomous EndoBot navigation under VISTA control in ex vivo flow experiments and an in vivo rat model.** (A) Autonomous closed-loop navigation in an ex vivo flow setup. (i–ii) Representative views of the integrated experimental platform at the start and end of navigation. The setup includes the GE Healthcare OEC 3D surgical imaging C-arm, the robotic magnetic actuation system, the vascular phantom, the blood-flow loop and pump system, the control computer, the live fluoroscopic image stream, and the Unity-based virtual environment. These panels illustrate the overall experimental configuration and the spatial relationship among imaging, magnetic actuation, and the target phantom. (iii–iv) Corresponding fluoroscopic and Unity-based views at the start and end of navigation. In the fluoroscopic images, the EndoBot and fiducial markers are visible, whereas the vessel segment remains radiolucent and is therefore not directly visualized. The detected 2D EndoBot position is mapped into the Unity environment, where the robot is localized relative to the reconstructed vessel centerline and used to estimate the navigation state for magnetic actuation control. (B) Autonomous closed-loop navigation in an in vivo rat model. (i) Representative view of the experimental setup, which includes the imaging and magnetic actuation platform together with a heat lamp used to maintain the animal's body temperature during the procedure. (ii) Corresponding fluoroscopic detection result and Unity-based virtual representation of the navigation environment. (iii–iv) Representative views of autonomous forward and backward navigation of the EndoBot within the inferior vena cava (IVC) under VISTA control.

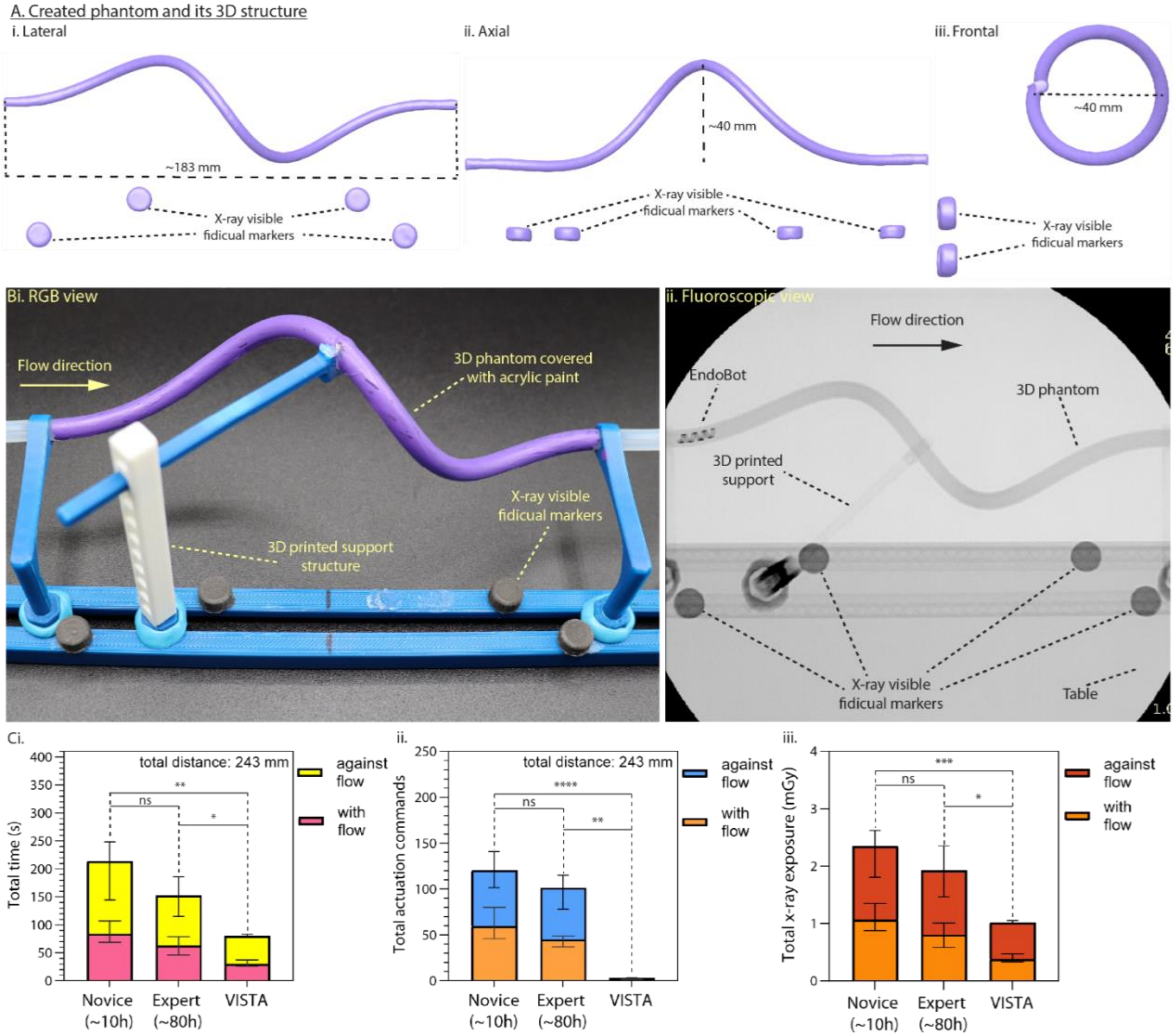
A. Created phantom and its 3D structure
i. Lateral
~183 mm
X-ray visible
fidicual markers
ii. Axial
~40 mm
X-ray visible
fidicual markers
iii. Frontal
~40 mm
X-ray visible
fidicual markers
Bi. RGB view
Flow direction
3D phantom covered
with acrylic paint
X-ray visible
fidicual markers
3D printed support
structure
ii. Fluoroscopic view
Flow direction
EndoBot
3D phantom
3D printed
support
X-ray visible
fidicual markers
Table
Ci.
total distance: 243 mm
against
flow
with
flow
Total time (s)
Novice
(~10h)
Expert
(~80h)
VISTA
ns
**
*
ii.
total distance: 243 mm
Total actuation commands
****
**
iii.
Total x-ray exposure (mGy)
***
*

**Figure 6 - Performance comparison of autonomous and manual navigation in a multiplanar helical vascular phantom.** (A) Design and fabrication of the vascular phantom used for performance evaluation. (i–iii) The custom phantom incorporated a 3D helical trajectory requiring navigation across multiple planes and was fabricated using a 3D-printed support structure and plastic tubing. The relevant geometric dimensions are indicated. (B) Experimental configuration for comparing VISTA with manual fluoroscopic control. (i–ii) The setup included the EndoBot, X-ray-visible fiducial markers, the external magnetic actuation system, a flow-pump circuit, and C-arm fluoroscopy. All control conditions were evaluated using the same phantom, imaging configuration, magnetic actuation platform, and flow conditions. (C) Quantitative comparison of navigation performance. VISTA, a novice operator, and an expert operator each completed three independent navigation trials. Performance was assessed using (i) total navigation time required for the EndoBot to complete the prescribed trajectory, (ii) total number of recorded robot-arm movement actions during navigation, and (iii) cumulative X-ray exposure.